\documentclass[lettersize,journal]{IEEEtran}
\usepackage{amsmath,amsfonts}
\usepackage{newfloat}
\usepackage{algorithmic}
\usepackage{algorithm}
\usepackage{array}
\usepackage[caption=false,font=normalsize,labelfont=sf,textfont=sf]{subfig}
\usepackage{textcomp}
\usepackage{stfloats}
\usepackage{url}
\usepackage{verbatim}
\usepackage{graphicx}
\usepackage{cite}
\usepackage{multirow}
\usepackage{amsthm}
\usepackage{amssymb}
\usepackage{hyperref}

\newtheorem{example}{Example}

\newtheorem{definition}{Definition}
\newtheorem{hypothesis}{Hypothesis}

\begin{document}

\title{Towards Causal Relationship in Indefinite Data: 
 Baseline Model and New Datasets}

\author{Hang Chen,~\IEEEmembership{Xinyu Yang}, Keqing Du

\thanks{The authors are with the Department of Computer Science and Technology, 
Xi'an Jiaotong University, Xi'an Shannxi, 710049.\\
E-mail: albert2123@stu.xjtu.edu.cn, yxyphd@mail.xjtu.edu.cn,\\dukeqing@stu.xjtu.edu.cn}}

\markboth{Journal of \LaTeX\ Class Files,~Vol.~14, No.~8, August~2021}%
{Shell \MakeLowercase{\textit{et al.}}: A Sample Article Using IEEEtran.cls for IEEE Journals}


\maketitle

\begin{abstract}
Integrating deep learning and causal discovery  
has encouraged us to spot that learning causal structures 
and representations in dialogue and video is full of challenges.  
We defined These data forms as ``Indefinite Data'', 
characterized by multi-structure data and multi-value 
representations. Unlike existing adaptable data forms, 
Indefinite Data still faces gaps in datasets and methods. 
To address the dataset gap, we release two high-quality 
datasets - \textit{Causalogue} and \textit{Causaction}, 
containing text dialogue samples and video action samples 
with causal annotations respectively. 
Moreover, the method gap arises from the coexistence 
of multi-structure data and multi-value representations, 
breaking the assumptions of all current methods, 
rendering them infeasible on Indefinite Data. 
To this end, we propose a probabilistic framework as a baseline, 
incorporating three designed highlights for this gap: 
1) establishing Causation Condition of representations 
using the independence of noise terms under 
non-fixed causal structures, 2) treating causal strength 
as a latent variable and measuring the reconstruction loss 
in the correlation space, and 3) estimating the effects 
of latent confounders. These highpoints 
make the probabilistic model capable of overcoming challenges 
brought by the coexistence of multi-structure data 
and multi-value representations, and pave the way 
for the extension of latent confounders. 
Comprehensive experiments have evaluated baseline results 
of causal structures, causal representations, 
and confounding disentanglement. Our codes are available at 
~\href{https://github.com/Zodiark-ch/master-of-paper-Towards-Causal-Relationship-in-Indefinite-Data-Baseline-Model-and-New-Datasets}{Github (click here)}. 
\end{abstract}

\begin{IEEEkeywords}
  Causal Data, Causal Representation, Causal Structures, Datasets, Baseline Model
\end{IEEEkeywords}

\section{Introduction}
In light of the recent advances in deep learning, 
there is a growing tendency to incorporate 
causal discovery in more complex forms of data, including 
images~\cite{jerzak2022image,ribeiro2023high}, 
text~\cite{zhang2023towards}, and videos~\cite{bagi2023generative}.
Generally, there are two purposes for these incorporations: 
one is to uncover the underlying \textbf{causal structure}~\cite{10148798,10214679,9805434,squires2022causal,zhang2017causal}
within the data, the other is to learn effective 
\textbf{causal representations}~\cite{9788000,204909,9956886,6701171,10146482,balashankar2021learning}. 

Our recent work~\cite{chen2023review} has summarized different 
forms of these incorporations  
based on causal structure and causal representation respectively. 
Regarding causal structure, there are \textbf{single-structure} data
~\cite{pearl2000models,yu2019dag,lachapelle2019gradient} 
and \textbf{multi-structure} data~\cite{lorch2022amortized,ke2020amortized,lowe2022amortized}, 
depending on whether multiple causal structures (causal graphs) 
are involved in the dataset or task. For example, 
fMRI dataset~\cite{smith2011network} suggests the different 
brain region activity levels of Patient $A$ and $B$, 
corresponding to two causal structures. Concerning causal 
representation, there are \textbf{single-value} representations~\cite{zhou2022causality,cai2019triad,9376668} 
and \textbf{multi-value} representations~\cite{li2021causality,9756301,oh2021causal}, 
depending on whether the causal variables need to be transformed 
into deep representations. Variables like age, height, weight, 
blood pressure are typically treated as single-value representations~\cite{647926}, 
while a sentence~\cite{chen2023learning} or a video~\cite{du2023casr} 
often needs to be converted by deep models into 
multi-value representations (such as sentence 
embeddings or optical flows) to make them calculable. 

Our work~\cite{chen2023review} further conjectured the emergence of a new 
causal data paradigm - \textbf{Indefinite Data} with 
the characteristics of both ~\textbf{multi-structure} data and 
~\textbf{multi-value} representations. For instance,  taking any 
dialogue as an input, could we recover the complete causal 
relationships between utterances and learn each utterance's 
causal representation? Or, if we replace the dialogue with a video, 
could we learn the internal relationships among segments 
and their corresponding causal representations? 

Despite the comprehensive definition provided by~\cite{chen2023review}, 
the study of Indefinite Data still faces two research gaps: 
the \textbf{dataset gap} and the \textbf{method gap}. 
Specifically, causal relationships in dialogues and videos 
are often obscure and subjective, making it challenging to 
collect samples with obvious causal relationships and 
objective annotations. Moreover, the co-occurrence 
of multi-structure data and multi-value representations 
breaks the hypotheses of all existing methods, 
resulting in their poor adaptability to Indefinite Data. 

To overcome these research gaps, we aim to release 
two high-quality Indefinite Datasets and a baseline model, 
specifically:

In Section 3, we analyze the causes of the dataset gap 
and particularly the method gap. Existing works on multi-value 
representations rely on a strong hypothesis that the 
causal structure is fixed and known~\cite{li2021causalhidden,xia2023deciphering,zhao2023generative}. 
Therefore, each causal variable can  
receive information from accurate parent set. Similarly, 
studies on multi-structure data operate under the hypothesis of 
single-value representations~\cite{varambally2023discovering,d2023learning,tran2023topological}, 
in which the precision of single-value representations 
provides indispensable statistical strength 
(e.g., reconstruction error~\cite{lorch2022amortized} or 
variations in distribution~\cite{lowe2022amortized}) for 
identifying multiple structures' invariances and dynamics. 
However, the co-occurrence of multi-value representations 
and multi-structure data breaks the fundamental hypotheses 
of these two mainstream methods, necessitating a 
redesign of how causal representations can be learned 
and how new causal structures can be adapted. 

In Section 4, we proposed a probabilistic framework 
as a baseline model based on Structural Causal Models (SCMs), 
featuring three novel designs: 1) The incorporation 
of an independent noise representation enables the 
output representation to discern specific causal relationships, 
thus completing the conversion from deep representations 
to causal representations. 
2) Treating the causal strength, rather than the noise term, 
as a latent variable avoids conflicts arising from different 
causal structure distributions. 3) The estimation of 
confounding effects disentangle the causal representation 
and the confounding representation, making the model enable to  
adapt to data with latent confounders. 

In Section 5, we introduced two brand-new Indefinite datasets - 
\textit{Causalogue} and \textit{Causaction}. 
\textit{Causalogue} is a text dataset containing dialogue samples 
used for analyzing causal relationships between utterances. 
To ensure the causal relationships are apparent and objective, 
we utilized GPT-4 to generate dialogues according to 
pre-defined causal rules. \textit{Causaction} is a video dataset 
containing different action segments, used for analyzing the 
causal relationships between different actions within a video. 
Annotators were asked to judge causal relationships 
directly based on low-level labels, rather than judging each 
video sample, thereby significantly reducing the 
subjectivity of causal relationships. 

In Section 6, we designed comprehensive evaluation metrics 
for Indefinite data on causal representation and causal structures, 
and compared them with some of the most adaptable SOTA methods. 
Additionally, to directly evaluate the performance of deconfounding, 
we also created a synthetic dataset with a known confounding 
distribution. 

In summary, for Indefinite Data, this paper provides 
why the gaps arise, what the baseline model looks like, 
how the high-quality datasets be created, 
and which evaluations should be concerned. 
Together with the basic definitions 
already proposed in our previous work~\cite{chen2023review}, 
it sets a promising onset for causal research in 
such causal data with fewer constraints and forms closer to 
the real world. 

\section{Preliminaries}

\begin{definition}[Causal representation]
  \label{defcr}
  The causal representation $\hat{X}$ represents the computed values 
  of causal variables when constructing a causal model. 
  Causal representations should meet the following two conditions:
  \begin{itemize}
    \item Correlation Condition: For any two causal variables 
    that exist correlation relationship, their corresponding causal 
    representations should contain the information of correlation. 
    \item Causation Condition: For any two causal variables 
    that exist causal relationship, 
    their causal representations should contain the  
    information about the causal relationship. 
  \end{itemize}
\end{definition}

For example, paper~\cite{fan2022debiasing} proposed that 
graph classification satisfies  
the SCM-based causal structure: $Y \leftarrow C\leftarrow G \rightarrow B$, 
where $G$ represents the observed graph, $C$ signifies 
the causal pattern, $B$ stands for the background pattern, and 
$Y$ represents the label.If $\hat{C}$, $\hat{G}$, and $\hat{B}$ 
correspond to the three causal representations of 
$C$, $G$, and $B$, 
there are no causal relationships but correlations between $C$ and $B$. 
As such,  the value of $cossim(\hat{C}, \hat{B})$ is close to $1$ 
(Correlation Condition). Furthermore, $\hat{C}$ and $\hat{B}$ 
should also meet the Causation Condition: 
for all samples satisfying this causal structure, 
the prediction results of the classifier should not change 
when $\hat{C}$ is combined with a set of  $\hat{B}$ from 
different samples.

We use $D$ to represent the dimensions of representation 
(i.e., $\hat{X} \in \mathbb{R}^{N\times D}$, where $N$ represents 
the number of causal variables) and there are two types of 
causal representations:

\textbf{Single-value Representation (D = 1)}: This type of variable 
inherently exists in numerical form, and thus, 
there is no necessity for the use of deep representation. 

\textbf{Multi-Value Representation (D $>$ 1)}: This type of variable 
doesn't inherently exist in numerical form and must be 
transformed into deep representations to enable computations.

  \begin{table}
    \caption{Examples about causal varaibles and representation}
    \label{tabecvcr}
    \resizebox{\linewidth}{!}{
    \begin{tabular}{|c|l|l|l|l|}
      \hline
      \textbf{Category}&\textbf{Variables}&\textbf{Deep model}&\textbf{Representation}&\textbf{Dimension}(D)\\
      \hline
      \multirow{2}{*}{Single-value}&Age &-& 25 (1-dimension value)&D=1\\
      &Voltage &-&2 (1-dimension value)&D=1\\
      \hline
      \multirow{2}{*}{Multi-value}&Token&RoBERTa&tensor&D=768, 1024\\
      &An image &LeNet-5&5*5-dimension tensor&D=5*5\\
      \hline
  \end{tabular}}
  \end{table}

  Table~\ref{tabecvcr} reveals the fundamental distinction 
  between single-value and multi-value representations. 
  Single-value representations are static, while multi-value 
  representations are dynamic (i.e., less precise). 
  Thus, single-value variables 
  often employ various statistical advantages to recover 
  causal structure, such as independence testing 
  and independent component analysis (ICA), while multi-value 
  representation can only rely on approximate correlation estimates, 
  such as similarity and divergence.

\begin{definition}[Causal structure]
  \label{defcs}
  The causal structure $\mathcal{G}$, represented as a causal graph 
  w.r.t. Directed Acyclic Graph (DAG), is used to describe a 
  set of causal relationships. 
\end{definition}

Many existing works\cite{lowe2022amortized,li2020causal,ke2020amortized} 
involved with multi-structure data  
demonstrate significantly different approaches 
compared to those associated with a single-structure data. 
Hence, we use $M$ to represent the number of structures and introduce 
two types of causal structures: 

\textbf{Single-structure Data (M = 1)}: For a given dataset or task, 
  there exists only a single causal graph $\mathcal{G}$, indicating 
  a fixed causal structure. 

\textbf{Multi-structure Data (M $>$ 1)}: For a given dataset, 
  multiple causal graph $\{\mathcal{G}\}^{M}_{m=1}$ exist, 
  implying an not unique causal strucutre for each sample. 

The difference between single-structure and multi-structure data 
lies in the fact that if a single-structure method 
is directly applied to multi-structured data, a new model needs 
to be refitted whenever a new causal structure is analyzed. 

Moreover, from above basic definitions, the definition of Indefinite 
Data is as following: 

\begin{definition}[Indefinite Data]
  The causal relationships exist in a dataset 
  $\mathbf{D} = \{X_{s}\}^{S}_{s=1}$ which has $S$ samples 
  and $M$ ($M > 1$) causal structures 
  ($\mathcal{G}=\{\mathcal{E}_{m}, \mathcal{V}_{m}\}^{M}_{m=1}$). 
  Each structure $\mathcal{G}_{m}$ corresponds to 
  several samples separately. 
  Hence, each sample $X_{s,m} \in \mathbb{R}^{N_{m} \times D}$  ($D > 1$)
  belongs to an individual causal structure  
  $\mathcal{G}_{m}=\{\mathcal{E}_{m}, \mathcal{V}_{m}\}$ and 
  consists of $N_{m}$ variables: 
  $X_{s}=\{x_{s,m,n}\}^{N_{m}}_{n_{m}=1}$. 
  $\hat{x}_{s,m,n} \in \mathbb{R}^{1 \times D}$ 
  represents the causal representation of a varaible $x_{s,m,n}$. 
  \label{defcausalmodel}
\end{definition}

\begin{example}[Indefinite Data]
  IEM Dataset~\cite{busso2008iemocap} is a conversation dataset 
  with each sample including a dialogue between two speakers. All 100 
  samples are assigned into 26 structures based on the 
  speaker identifies and turns. Each sample consists of 5-24 
  causal variables where each variable is an 
  utterance represented by word embeddings. 
  \label{expid}
\end{example} 

Moreover, given the major examples of Indefinite Data 
involves textual conversations and video sources, 
we propose a hypothesis about the causal identifiability: 

 \begin{hypothesis}[Causal Identifiability]
   The natural order (e.g., time-order) \textit{w.r.t.} 
   $\{x_{s,m,n}\}^{N_{m}}_{n_{m}=1}$ 
   is defined as a linear order $\prec_{X_{s,m}}$. Given that 
   causal order \textit{w.r.t.} $\{x_{s,m,n}\}^{N_{m}}_{n_{m}=1}$ 
   is defined as a partial order $\preccurlyeq_{X_{s,m}}$, 
   $\forall <x_{1},x_{2}> \in \prec_{X_{s,m}} (i.e., x_{1} \prec_{X_{s,m}} x_{2})$, 
   there must be $<x_{1},x_{2}> \in \preccurlyeq_{X_{s,m}}$. 
   \label{hyp1}
   \end{hypothesis}

Hypothesis~\ref{hyp1} illustrates the natural linear order 
of Indefinite data (e.g., $\{U_1, U_2, U_3, U_4\}$, 
where $U_1$ to $U_4$ respectively represent 4 utterances 
appearing in time-series, and $U_i\prec U_j$ indicates that 
$U_i$ precedes $U_j$ in time) belongs to the causal partial order. 
Consequently, the adjacency matrix of the natural linear order 
is a triangular matrix, which naturally corresponds to a DAG. 
Thus, there is no need for measures such as acyclic constraints
~\cite{zheng2018dags} to ensure causal identifiability.

\section{Research Gaps of Indefinite Data}

\subsection{Related Work}

\subsubsection{Multi-value Representation $\&$ Single-structure Data}
In the domain of images, \cite{lopez2017discovering} initially 
treats object features and context features as two 
causal representations, learning via the integration of 
additive noise models (ANMs) and neural networks. Subsequent works 
followed this pathway and developed various methods to 
extract causal representations, such as LSTM~\cite{oh2021causal} 
and linear layers~\cite{shadaydeh2021analyzing,mao2022causal}, 
along with more refined causal variables~\cite{li2021causalhidden,xia2023deciphering,zhao2023generative}. 

As for text, existing studies, based on prior knowledge, 
pre-set that certain words carry essential 
causal clues or interferences at the word-level embedding, 
such as verbs~\cite{beltagy2019scibert}, conjunctions~\cite{zhao2016event}, 
and terminologies~\cite{khoo2000extracting}. 
For utterance-level embeddings, the SCM is often used as 
guiding prior knowledge, spurring a vast amount of work 
on the generation of exogenous latent causes~\cite{chen2023affective,chen2023learning}.

In other fields, like audio or graph, 
decoupling at the representation level often occurs~\cite{wu2021discovering,hazan2007causal,fan2022debiasing}. 
This results in splitting the representation into a set of 
variables that have causal relationships with labels, 
and another series of variables exhibiting spurious correlation 
with labels. 

\subsubsection{Multi-structure Data $\&$ Single-value Representation}
A main body of work refers to methods for addressing such data 
as amortized learning or mixed models learning~\cite{lowe2022amortized,ke2020amortized,lorch2022amortized}, 
exploring the linear mixed effects models~\cite{li2018learning}, 
multiple amortized structures~\cite{ke2020amortized,li2020causal}, 
and across samples learning~\cite{dhir2020integrating,huang2020causalb,huang2020causala}. 
The range of approaches to multi-structure data relevant to 
single-value representation including~\cite{lowe2022amortized} who 
utilizes the reconstruction error $loss(X,\hat{X})$ to control the 
distribution of causal strength, concurrent work find the invariable 
causal relationship across structures via statistics of single-value 
representation~\cite{li2020causal,li2018learning}.

\subsection{Method Gap}

Indefinite Data (D $>$ 1 $\&$ M $>$1) can be simply considered 
as an integration of two types of  D = 1 $\&$ M $>$ 1 and 
M = 1 $\&$ D $>$ 1. However, 
the two types both rely on the hypothesis that the other 
dimension is $=1$. 

When we aim to design a model to learn multi-value causal 
representations, the infer process always inspired by fixed, 
prior-regarded causal structures. 
Formally, we assume $\hat{X} = infer(X)$, where $infer(\cdot)$ 
is designed through a fixed causal structure. 
For instance, 
~\cite{fan2022debiasing} believe that the background $B$ of a 
graph $G$ could mislead the labels $Y$ caused by causal pattern $C$,  
due to a fork structure on the path between $B$ and $Y$, 
as $Y \leftarrow C\leftarrow G \rightarrow B$. 
the Causation Condition of the causal representation 
should be satisfied as: $\hat{X}=infer(P(Y|do(C)))$. 
It describes that the strength of the 
causal pattern $C$ to the $Y$ should remain unchanged
with replacing any $B$ from other samples. 
Alternatively, when we discover that spatial features have created  
front-door paths to the label, the Causation Condition should 
be equal to the what an intermediate variable $X^{*}$ satisfies~\cite{xia2023deciphering}, 
which reads: $\hat{X}=infer(P(Y|X^{*}))$. 
In summary, if we break the hypothesis of single-structure data, 
there is a lack of causal clue to formulate the Causation Condition. 

Similarly, when we wish to learn some invariants  
from multi-structure data, 
some unchanged causal relationships can be uncovered 
from single-value representations. 
That is, $G=infer(X)$, where $infer(\cdot)$ is designed through 
single-value representation. For example, ~\cite{li2020causal} 
proposed $G=infer(\{p(X_m)\}^{M}_{m=1})$ to decouple 
different structural distributions relying on the accurate statistics 
value of single-value representation. 
Furthermore, the reconstruction loss of a 
single-value representation, written as $G=infer(loss_{rc}(X,\hat{X}))$, 
also theoretically supports the ELBO of the 
posterior distribution of causal strengths~\cite{lowe2022amortized}. 
This is not achievable in multi-value representations, 
as $X$ only satisfies the Correlation Condition, while $\hat{X}$ 
satisfies the Causation Condition. 

Taking deep model as an instance, $p_{\varphi}$ and 
$q_{\theta}$ represents the encoder and decoder of generative model 
from domain of causal variable $X$ to the domain of 
causal representation $\hat{X}$ where 
$f_{i}$ is causal strength responsible for the causal mechanism. 

\textbf{ M = 1 $\&$ D = 1}: 
The causal strength can be estimated by the statistical strength 
observable in the samples. 

\textbf{M $>$ 1 $\&$ D = 1}: 
We can separate the problem to several tasks of 
single-structure data. Reconstruction loss amounts to 
$\{f_{m}\}^{M}_{m=1}$, 
where can be regarded as a multi-task optimization problem, 
$\alpha_1 f_1 + \alpha_2 f_2 +\dots+\alpha_M f_M$, where $\alpha_m$ 
is the weights of the sample quantity per structure. 

\textbf{M = 1 $\&$ D $>$ 1}: 
The reconstruction loss can be written as: 
$p_{\varphi}\circ f\circ q_{\theta}$, where $f$ represents the 
determined part due to fixed causal structure. 

\textbf{M $>$ 1 $\&$ D $>$ 1}: 
We are only able to attain an approximate 
$\tilde{p}_{\varphi}=p_{\varphi}\circ f_m$, 
which results in a final reconstruction loss of 
$\tilde{p}_{\varphi} \circ q_{\theta}$. Causal strength $f_m$ 
comprises an undetermined part. 

\subsection{Dataset Gap}

In our previous work~\cite{chen2023review}, we collected a set 
of public datasets satisfying the requirements of Indefinite Data. 
The limitations mainly arise from the fact that Indefinite Data 
largely exists in continuous forms, making the demarcation 
of causal variable boundaries a significant challenge. 
For instance, in the arithmetic datasets~\cite{roy2016solving,talmor2018commonsenseqa,wei2022chain}, 
although much effort has been devoted to discover 
the relationships between different steps, 
the non-uniqueness of the reasoning process makes it difficult 
to transcribe the chains of thought (CoT) into a set of 
causal variables. However, video and dialogue datasets 
do have clear variable boundaries; for example, 
video datasets~\cite{lee2019context,stein2013combining} 
can be segmented based on action semantics, and each utterance 
in dialogue datasets is discrete~\cite{li2017dailydialog,busso2008iemocap}. 
However, on these datasets, most causal relationships are obscure, 
which leads to poor consistency in manual annotation. 
Taking dialogue as an example, utterances 
that have not been observed before might likely act 
as confounding factors influencing the correlation 
between observed utterances. Moreover, 
the standards for judging whether there is a causal relationship 
between two utterances is terribly subjective. Up to now, 
only a fraction of the work~\cite{poria2021recognizing,chen2023affective} 
has annotated some evident causal 
relationships, and no complete causal-labeled dataset 
has yet appeared, which significantly dampens researchers' 
enthusiasm for Indefinite Data. 

\section{Baseline Model}

\subsection{Fundamental Framework}
\label{fundamentalfw}

Considering the latent confounders, the SCM is written as: 

\begin{equation} \
  \begin{split}
    x_{m,j}=&\sum_{x_{m,i}\in Pa(x_{m,j})} f_{m,ij} x_{m,i}\\
    &+\sum_{l_{m,k}\in Ec(x_{m,j})} g_{m,kj} l_{m,k} +\epsilon_{x_{m,j}}
    \label{eqt17}
  \end{split}
  \end{equation}
  where $Pa(x_{m,j})$ represents the parent set of $x_{m,j}$, 
  $Ec(x_{m,j})$ is the confounder set having effects on $x_{m,j}$, 
  $f$ and $g$ denotes the causal strengths and confounding strength, 
  respectively, $\epsilon$ represents 
  the exogenous i.i.d., noise term, and $K$ is the number of latent confounders. 
  $x \in \mathbb{R}^{N \times D},l \in \mathbb{R}^{K \times D},\epsilon_{x_{s,j}} \in \mathbb{R}^{N \times D}, 0\leq i \neq j < N, 0\leq k < K$. 
  The matrix form reads: 
  \begin{equation}
   X=AX+BL+E
   \label{eqt19}
  \end{equation}
  
We design a couple of encoder and decoder to model the 
generating process of causal representation: 

\begin{align}
  Encoder:& W=f_{1}(X)\\
  Decoder:& \hat{X}^{*} =f_{2}(W(BL+E))
  \label{eqt29}
\end{align}
where $f(\cdot)$ perform nonlinear transforms (neural network as 
GNN or MLP layers are popular choices) and $W$ represent $(I-A)^{-1}$. 
Please note, $f_{1}(X)$ is an abbreviation of $f_{1}(X(BL+E)^{-1})$ 
as $X$ consist of $BL+E$ w.r.t. $W$. 
Decoder can be written by a maximization of leg-evidence: 

\begin{equation}
\begin{split}
  &\frac{1}{M}\frac{1}{S} \sum_{m=1}^{M}\sum_{s=1}^{S} \log p(X_{s,m})=\\
  &\frac{1}{M}\frac{1}{S} \sum_{m=1}^{M}\sum_{s=1}^{S} \log\int p(X_{s,m}|W)p(W)dW
  \label{eqt15}
\end{split}
\end{equation}

Continuing the theory of variational Bayes, we regard $W$ as 
the latent variable in variational autoencoder (VAE)
~\cite{kingma2022autoencoding} and use variational 
posterior $q(W|X)$ to approximate the intractable posterior 
$p(W|X)$, thus the evidence lower bound (ELBO) reads: 

\begin{equation}
  \begin{split}
    \mathcal{L}^{s,m}_{ELBO}=&-KL(q(W|X_{s,m})||p(W))\\
    &+E_{q(W|X_{s,m})}[\log p(X_{s,m}|W)] 
    \label{eqt16}
  \end{split}
  \end{equation}
For simplicity, we model the prior as the standard normal 
$p(W)=\mathcal{M}\mathcal{N}_{N \times N}(0,I,I)$, which indicates 
that each causal strength $p(f_{ij})=\mathcal{N}(0,1)$. Note that 
even though the nodes are probably connected in a true graph, however, 
they are independent in prior. 

In the causal view, our framework consists of two functions: a causal 
strength encoder: $\mathcal{X} \rightarrow \mathcal{G}$ and a 
causal representation decoder: 
$\mathcal{G} \rightarrow \widehat{\mathcal{X}}$.

\subsection{Estimation of Confounding Effect}\label{eoce} 

We use $c_{m,j}=\sum_{l_{m,k}} g_{m,kj} l_{m,k}$ 
to describe the confounding effect on $\widehat{x}_{m,j}$ 
and $C=BL$ to describe the corresponding matrix form. 
Inspired by~\cite{agrawal2021decamfounder}, we proposed a 
estimation about $C$: 

\begin{equation}
  c_{m,j}=\frac{p(x_{m,j}) p(L|x_{m,j})}{\sum_{i}^{N} p(x_{m,i}) p(L|x_{m,i}) } x_{m,j} 
  \label{eqt26}
\end{equation}

Equation~\ref{eqt26} only works when the expectation 
$E_{p(X)}(X|L)$ is much greater than the expectation $E_{p(X)}(X|\epsilon)$. 
It collaborates the inductive bias that 
when confounding effects drastically exceed independent noise, 
$X$ is approximately contributed by $C$ rather than $E$. 
Therefore, the disentangled causal representation $\hat{X}$ 
can be written as: 

\begin{equation}
  \begin{split}
    \hat{X}\approx \left\{
      \begin{array}{lr}
        \hat{X}^{*}-C, &(E_{p(X)}(X|L) \gg E_{p(X)}(X|\epsilon))\\
        \hat{X}^{*}, &else
      \end{array}
    \right.
  \end{split}
  \label{eqt265}
\end{equation}

\subsection{Explanation}
\subsubsection{How to Extend to Multi-value Representation?}
Given the existence of deconfoundment, we can, 
without loss of generality, write the SCM as: 
$\hat{x_{j}}=\sum_{\hat{x_{i}}\in Pa(x_{j})} f_{ij} \hat{x_{i}}+\epsilon_{x_{j}}$, 
where the independence of $\epsilon$ ensures the Causation Condition. 
That is, we can directly recover the causal relationship from the 
causal representation $\hat{x}$. For instance, 
if we linearly make causal representations 
$\hat{a}$ to fit $\hat{b}$ with an learnable parameter $k$ in a 
downstream task, and obtain the corresponding residuals: 
$\Sigma_{\hat{b}}=\hat{b}-k\hat{a}, \Sigma_{\hat{a}}=\hat{a}-\frac{1}{k}\hat{b}$. 
Then, different causal relations can be determined through 
the independence combination between residuals and representations: 

\begin{itemize}
  \item $\Sigma_{\hat{a}} \perp \!\!\! \perp \hat{b}, 
  \Sigma_{\hat{b}} \not \! \perp \!\!\! \perp \hat{a}
  \Rightarrow \hat{b} \rightarrow \hat{a}$
  \item $\Sigma_{\hat{a}} \not \! \perp \!\!\! \perp \hat{b}, 
  \Sigma_{\hat{b}} \perp \!\!\! \perp \hat{a}
  \Rightarrow \hat{a} \rightarrow \hat{b}$
  \item $\Sigma_{\hat{a}} \not \! \perp \!\!\! \perp \hat{b}, 
  \Sigma_{\hat{b}} \not \! \perp \!\!\! \perp \hat{a}
  \Rightarrow l \rightarrow \hat{a}, l \rightarrow \hat{b}$
  \item $\Sigma_{\hat{a}} \perp \!\!\! \perp \hat{b}, 
  \Sigma_{\hat{b}} \perp \!\!\! \perp \hat{a}
  \Rightarrow \hat{a} \rightarrow l, \hat{b} \rightarrow l$
\end{itemize}

\subsubsection{How to Extend to Multi-structure Data?}
In contrast with popular methods that intuitively treat the 
noise matrix as a latent variable~\cite{yu2019dag,chen2023affective} 
($\mathcal{X} \rightarrow \mathcal{E}$ and $\mathcal{E} \rightarrow \widehat{\mathcal{X}}$), 
we attempt to regard the causal strength as a latent variable, 
thereby enabling one model to learn multiple structures. 
From the overall view, sampling from a set 
of DAGs $\mathcal{G}_{m}=\{\mathcal{E}_{m}, \mathcal{V}_{m}\}^{M}_{m=1}$ 
is equal to generate a set of causal strengths which reads: 
\begin{equation}
  p(A)=\{p(A_m)\}^{M}_{m=1}
\label{eqt12}
\end{equation} 

Moreover, to overcome the limitation that it is impossible 
to construct a loss function $loss(x, \hat{x})$ for 
multi-value representation, we map $x$ and $\hat{x}$ 
onto the space of Correlation relationships 
(see equation~\ref{eqt34} for details). 
That is, $x$ and $\hat{x}$ remain consistent in the 
Correlation Condition, they conflict in the 
Causation Condition though.

\subsection{Implementation Example}

We formalized the dynamic variational inference model as follows: 
a causal strength encoder 
$f_{\varphi}: \mathcal{X} \to \mathcal{G}$, an causal representation 
decoder $f_{\theta}:\mathcal{G} \to \widehat{\mathcal{X}}$, 
and an estimation function 
$f_{\delta}:\mathcal{X} \to \mathcal{C} $. 

\begin{figure*}
  \includegraphics[width=1\linewidth]{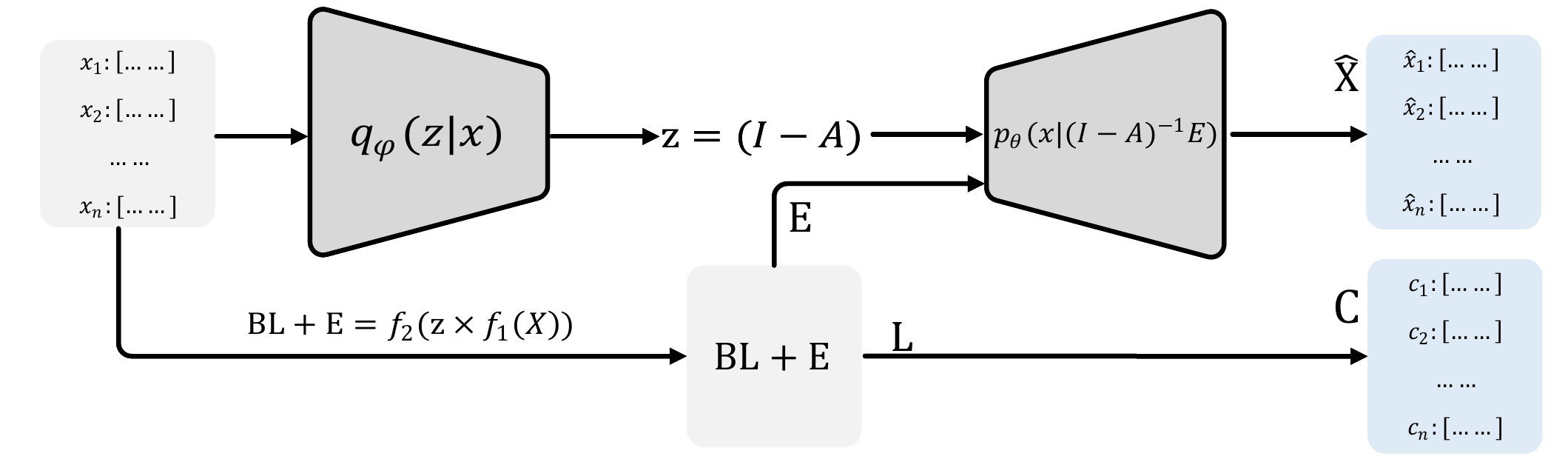}
  \caption{An implementation example of our framework. 
  $q_{\varphi}(z|\mathcal{X})$ predicts the causal strength 
  from the input $X$. The predicted latent variable $z=(I-A)$, 
  and then an causal representation decoder 
  $p_{\theta}((x|(I-A)^{-1}E))$ learns to predict 
  $\widehat{X}$ given the disentangled $E$ 
  and inverse of predicted $z$.}
  \label{figbiCD}
\end{figure*}

We resort to VAE to design the functions $f_{\varphi}$ 
and $f_{\theta}$ as shown in Figure~\ref{figbiCD}. Specifically, 

\subsubsection{Encoder}
The encoder $q_{\varphi}(z|\mathcal{X})$ applies a graph attention 
module $f_{att,\varphi}$~\cite{velivckovic2017graph} to the input. 
It produces an adjacent matrix across a lower triangular mask 
under Hypothesis~\ref{hyp1}. 
\begin{equation}
q_{\varphi}(z|\mathcal{X})=softmax(f_{att,\varphi}(X))
\label{eqt27}
\end{equation}

The output $z$ implies the possible distribution of causal strength 
over $\mathcal{X}$. Specifically, $z_{i,j}=1$ indicates a 
high probability relation $x_{j} \to x_{i}$. 

\subsubsection{Decoder}

we extract $E$ by utilizing a multi-layer 
perceptron (MLP): 
\begin{align}
  BL+E=&GNN_{enc}(f_{att,\varphi}(X), X)\\
  E = &MLP_{E}(BL+E)
  \label{eqt29}
  \end{align}
where $GNN_{enc}$ is instantiated by graph neural network: 
$GNN(\mathsf{A}  ,\mathsf{X} )=eLU(\mathsf{A} \times (\mathsf{X} \times \mathsf{W}) )$, 
which yields a nonlinear multiple of adjacent matrix $\mathsf{A} \in \mathbb{R}^{N \times N}$, 
feature matrix $\mathsf{X}  \in \mathbb{R}^{N \times D}$ 
and weight matrix $\mathsf{W}  \in \mathbb{R}^{D \times H}$, 
where $H$ represents the dimensions of hidden layers. 
Then, the decoder accumulated the incoming messages to each node via 
causal strength $z$ and employed a new graph neural network $GNN_{dec}$: 
  \begin{align}
    p_{\theta}((\hat{x}_c|z^{-1}E))=GNN_{dec}(z^{-1},E)
    \label{eqt30}
    \end{align}
The output of the decoder $\hat{x}_c\in \mathbb{R}^{N \times D}$ equals 
the dimension of $\mathcal{X}$ and  it is the pure causal representation 
of $\hat{x}$ without confounding. 

\subsubsection{Confounding Estimation}
We used the same MLP module to extract $L$ and two sigmoid functions: 
$\sigma_{p(x_{j})}(\cdot )$ and $\sigma_{p(L|x_{j})}(\cdot )$, 
to project $p(x_{j})$ and $p(L|x_{j})$ into the range 
of $(0,1)$, which expresses the probability estimating $c_{j}$. 
\begin{align}
  &L = MLP_{L}(BL+E)\\
  &c_{j}=\frac{\sigma_{p(x_{j})}(x_{j} ) \sigma_{p(L|x_{j})}(L|x_{j})}{\sum_{i}^{N} \sigma_{p(x_{i})}(x_{i} ) \sigma_{p(L|x_{i})}(L|x_{i}) } x_{j} 
  \label{eqt332}
  \end{align}
The output $C \in \mathbb{R}^{N \times D}$ of Estimation module equals 
the individual-specific effects of confounding on each $x_{j}$ if there 
exactly exists strong confounding. 

\subsubsection{Reconstruction Error}

Given the contradiction between $X$ and $\hat{X}$ 
on the Causation Condition explained in Definition~\ref{defcr}, 
we map the $X$ and $\hat{X}$ into the correlation space. 

Moreover, 
considering the dynamics of confounding effects across samples 
(as shown in Equation~\ref{eqt265}), 
we naturally design a confounding score for each graph as 
$\omega(L)=rank(L)/N$ ($L$ is computed by equation~\ref{eqt332}. 
The graphs with high $\omega(L)$ can be regarded as confounding 
samples because the high rank of the matrix $L\in \mathbb{R}^{N \times D}$ 
stands for the extensive independent terms in $L$, which indicates 
that sufficient exogenous confounding variables point to the $X$, and 
vice versa. Finally, the reconstruction error and ELBO can be encapsulated by: 
\begin{align}
  &l_{RC}=\omega(L) l_{rc}(X,\widehat{X}+C)+ (1-\omega(L))l_{rc}(X,\widehat{X})\\ 
  &\mathcal{L}=l_{RC}-KL[q_{\varphi}(z|\mathcal{X})||p(z)]
  \label{eqt40}
  \end{align} 

Specifically, We adopt mean squared error (MSE) and 
cosine similarity in implementation: 
\begin{align}
  l_{rc}(X,\widehat{X})=&E_{q_{\varphi}(z|\mathcal{X})}[MSE(cs(X),cs(\hat{X})]\\
  l_{rc}(X,\widehat{X}+C)=&E_{q_{\varphi}(z|\mathcal{X})}[MSE(cs(X),cs(\hat{X}+C)]\\
  cs(\cdot)=&\sum_{0\leq i < j < N} cossim(\cdot_{i},\cdot_{j})\\
  \label{eqt34}
  \end{align}

\section{New Datasets}
\subsection{\textit{Causalogue}}

\subsubsection{Attributes}
\textit{Causalogue} is the first dialogue dataset that 
includes comprehensive causal relationship labels 
for Indefinite data. Additionally, we employ GPT-4 generation 
as a substitute for data collection from the real world or 
manual simulation, which considerably mitigates 
the presence of obscure causal structures. 

The dataset incorporates 10 types of causal structures (M = 10), 
each with several samples 
(Detailed numbers are presented in Table~\ref{tabscd}, 
``Small" signifies samples that have been manually checked, 
while ``large" refers to all samples 
generated by GPT-4 without manual verification). 
The detailed attributes are following:

\textbf{Causal Variables}: We treat each dialogue as a sample, 
comprised of 4 utterances, 
which we define as 4 causal variables. 
Further, the first and third utterances originate from 
the same speaker, defined as $speaker 1$. 
Similarly, the second and fourth utterances are 
from another individual, defined as $speaker 2$. 

\textbf{Causal Relationship}:In each sample, binary causal relationships 
have been labeled between any two utterances-``1''represents that 
there exists a causal relationship and ``0'' represents there not. 

\begin{figure*}
   \centering
   \subfloat[Chain$\_$I]{
     \includegraphics[width=0.16\textwidth]{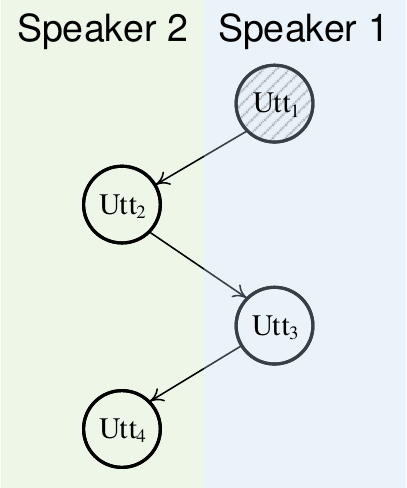}}
   \subfloat[Chain$\_$II]{
     \includegraphics[width=0.16\textwidth]{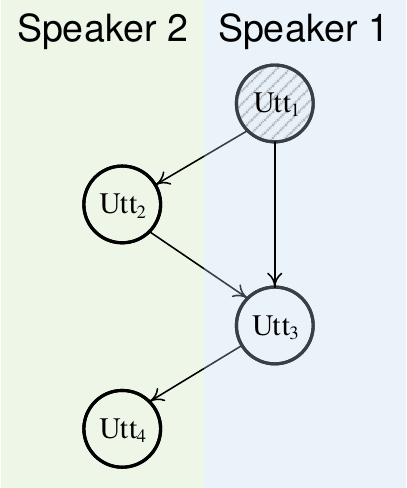}}
   \subfloat[Chain$\_$III]{
     \includegraphics[width=0.16\textwidth]{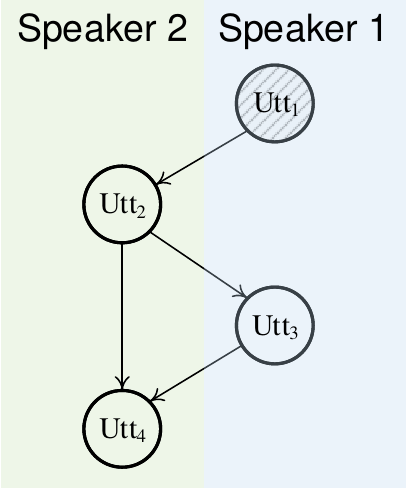}}
   \subfloat[Chain$\_$IV]{
     \includegraphics[width=0.16\textwidth]{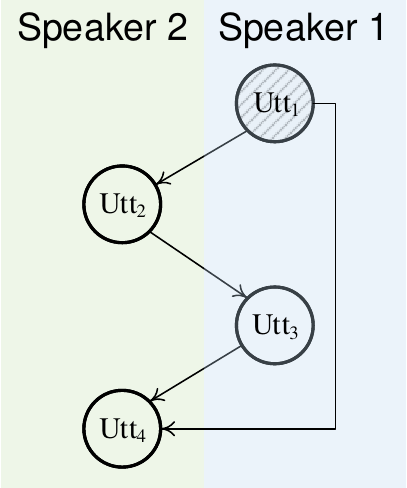}}
   \subfloat[Hybrid$\_$I]{
     \includegraphics[width=0.16\textwidth]{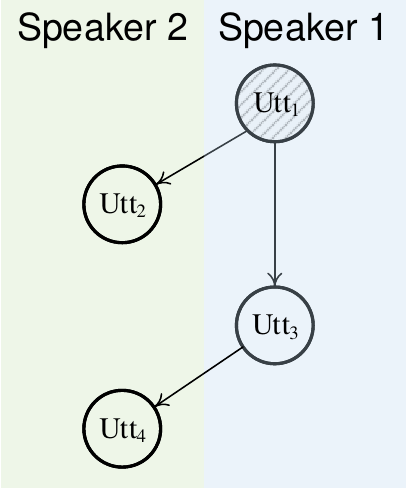}}\\
   \subfloat[Fork$\_$I]{
     \includegraphics[width=0.16\textwidth]{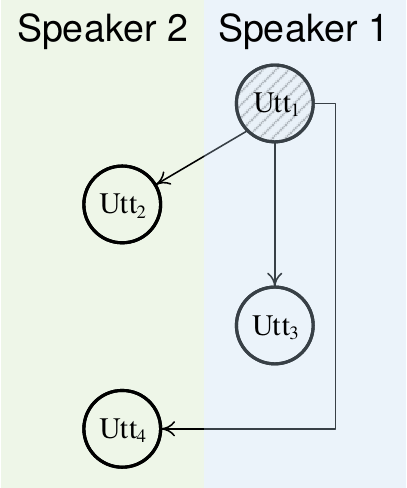}}
   \subfloat[Fork$\_$II]{
     \includegraphics[width=0.16\textwidth]{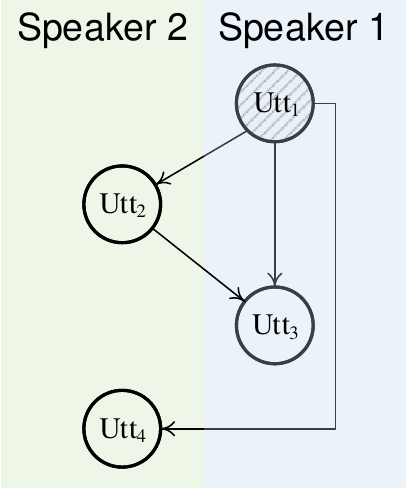}}
   \subfloat[Fork$\_$III]{
     \includegraphics[width=0.16\textwidth]{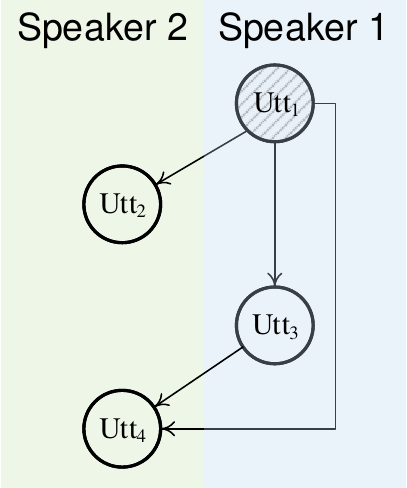}}
   \subfloat[Fork$\_$IV]{
     \includegraphics[width=0.16\textwidth]{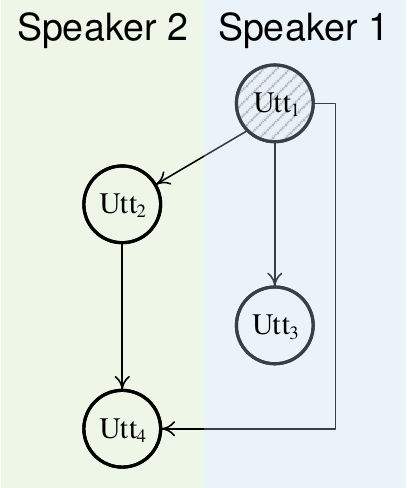}}
   \subfloat[Hybrid$\_$II]{
     \includegraphics[width=0.16\textwidth]{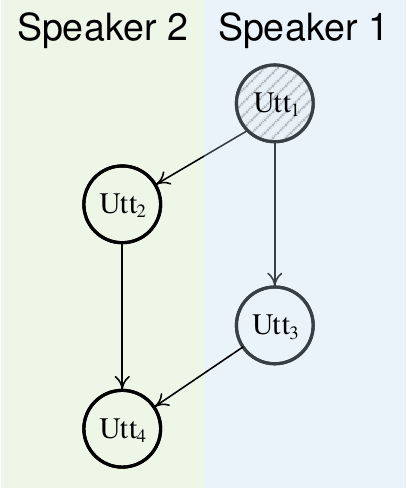}}
   
   \caption{10 structures in \textit{Causalogue} Dataset.}
   \label{fig10s}
 \end{figure*}

\textbf{Structure}: We have designed 10 types of causal structures 
in the dataset as shown in Figure~\ref{fig10s}. Taking the Chain$\_$II 
as an example,  this model adds  
an additional causal relationship from $Utt_1 \rightarrow Utt_3$ 
based on the Chain$\_$I, 
indicating that $Utt_3$ considers not just the effect  
from $Utt_2$ but also from $Utt_1$.

\textbf{Sample}: We consider a dialogue as a sample, 
with each sample comprising 4 utterances representing 
4 causal variables. Each sample corresponds to one of the 
10 causal structures outlined above, 
annotating whether a causal relationship exists 
between any two utterances. Due to Hypothesis~\ref{hyp1}, 
our labels only consider forward-causal relationships. 
An example of a Chain$\_$III sample is shown as follows:

\textit{``causal$\_$type'': ``Chain$\_$III'',}

\textit{``clause'': $\{$``1'': ``Your bill is 19.'', 
``2'': ``Before I pay the bill, I have to express 
my dissatisfaction with the service I received tonight.'', 
``3'': ``I'm so sorry to hear that but I don't know what happened.'', 
``4'': ``Specifically, It's understandable to 
feel frustrated when something unexpected happens 
like spilling red wine on your clothes.''$\}$, }

\textit{``dia$\_$id'': 1,}

\textit{``label'': $\{$``1'': ``0,0,0,0'', ``2'': ``1,0,0,0'', 
``3'': ``0,1,0,0'', ``4'': ``0,1,1,0''$\}$}

In the given example, the $Utt_4$ serves as a response 
to the $Utt_3$, while simultaneously attach to the speaker's 
$Utt_2$—thereby rendering both the $Utt_2$ and $Utt_3$ 
as causes to the $Utt_4$. Indeed, during the generation process 
of the $Utt_4$, we made sure to inform GPT-4 of 
the existence of $Utt_2$ and $Utt_3$.

\begin{table*}
   
  \caption{Number of the samples in \textit{Causalogue} Dataset}
  \label{tabscd}
  \begin{center}
     \resizebox{1\linewidth}{!}{
  \begin{tabular}{c|ccccccccccc}
  \hline
  \multirow{2}{*}{\bf Versions}&\multicolumn{11}{c}{\bf Structure Types}\\
  \cline{2-12}
  &Chain$\_$I&Chain$\_$II&Chain$\_$III&Chain$\_$IV&Fork$\_$I&Fork$\_$II&Fork$\_$III&Fork$\_$IV&Hybrid$\_$I&Hybrid$\_$II&Total\\
  \hline
  Small&276&84&141&44&257&237&251&67&185&77&1638\\
  \hline        
  Large&0&524&508&513&1215&645&501&372&499&635&5412\\
  
  \hline
  \end{tabular}}
  \end{center}
  \end{table*}

\subsubsection{Creation Process}

We utilized the API interface of GPT-4~\footnote{\url
{https://platform.openai.com/docs/models/gpt-4}} 
to defined the following variables: ``\textit{role}'', 
which has three types - ``\textit{system}'', 
``\textit{user}'', and ``\textit{assistant}''. 
Here, ``\textit{system}'' represents the background 
or a prior settings, while ``\textit{user}'' and 
``\textit{assistant}'' are defined as speakers 
with two different identities. Additionally, 
the first utterance is pre-set. Hence, 
creating a dialogue requires a given combination: 
a fixed \textit{first$\_$utterance}, 
a specified \textit{system} information, 
and a setting which previous utterances are considered. 
We have a total of 149 \textit{first$\_$utterance} options, 
and there are as many as 278,867 combinations of 
\textit{first$\_$utterance} and \textit{system} settings 
(our final samples only number in the 1638, 
to preserve the diversity and distinctiveness of our dialogues). 
What follows is an example of generating the third utterance 
in the structure of Chain$\-$II:

\textit{$\{$``role'': ``system'', ``content'': ``You are Peter, 
   you have promised to go to a Chinese Opera with your daughter, 
   so you want to have dinner with your 
   friends in next Sunday.'' $\}$}
   
   \textit{$\{$``role": ``assistant'', ``content'': ``Yes. Sunday sounds
    fine. What time?'' (pre-set Utt$\_$1)$\}$}
   
    \textit{$\{$``role'': ``user'', ``content'': Utt$\_$2$\}$}

Upon creation, the samples are initially auto-annotated 
based on their designed labels, and then manually verified 
to ensure their validity. Our manual verification employed 
two annotators, who demonstrated proficient English understanding 
and communication skills, possessing 
sufficient knowledge about causality. The annotation consistency  
between these two annotators was tested through 833 samples, 
achieving a kappa coefficient of 0.92. 

During the annotation process, if a sample was labelled 
differently by the two annotators, that sample was considered 
to possess an ambiguous causal relationship and thus 
was excluded from the final dataset. 
Only samples that were consistently labeled by both annotators 
were ultimately accepted. 

Furthermore, to guarantee the freedom 
of manual annotation, we allowed the annotators to label 
structures that fell outside the predefined 10 causal structures. 
Specifically, we only requested annotators to judge whether 
any two utterances (satisfying Hypothesis~\ref{hyp1}) 
have a causal relationship, allowing them some discretion, 
which inevitably produced samples 
not belonging to the 10 causal structures. 
We classified these as the ``Other'' category. 

The accuracy of labels was significantly improved 
after the manual annotation process. However, considering 
that the unverified samples might be utilized for other research 
areas, such as the ability of LLMs to focus on context, 
we have released two versions of the datasets, 
as demonstrated in Table~\ref{tabscd}. 
``Small" signifies samples that have been manually checked 
as correctly labeled, while ``large"refers to all samples 
generated by GPT-4 without manual verification. 
We do not recommend considering the ``large" version 
when undertaking causality-related work. Likewise, 
we have not taken it into our experiments.

\subsection{\textit{Causaction}}
\subsubsection{Attributes}
\textit{Causaction} is another Indefinite Dataset 
that we obtained after re-annotating the Breakfast Dataset~\cite{kuehne2014language}. 
It contains a total of 1,118 videos, documenting 10 
different breakfast preparation processes  
(such as coffee, salad, sandwich, etc.). Each video consists of 
4-9 actions, with a clear frame boundary. 
We have annotated the causal relationship 
between any two actions in a sample. 
Specific attributes are as follows: 

\textbf{Causal Variable}: We treat each video as a sample, comprised 
of 4-9 actions as the causal variables. For simplicity, 
we follow the setting of MS-TCN~\cite{farha2019ms}, replacing the video 
resource of each action with pre-trained representation of I3D~\cite{carreira2017quo}.

\textbf{Causal Relationship}: According to 
the Hypothesis~\ref{hyp1}, we deem the time order of these actions 
in certain video as natural linear order. Hence, binary causal 
relationships have been labeled between any two actions satisfying 
the linear order (`0' represents there is no causal relationship while 
`1' represents there is). For example, process ``cereals'' includes 
4 actions: ``take bowl'', ``pour cereals'', ``pour milk'', and ``stir cereals''. 
The all causal relationships labeled with ``1'' are: 
``take bowl $\rightarrow$ pour cereals'', ``take bowl $\rightarrow$ pour milk'', 
``take bowl $\rightarrow$ stir cereals'', ``pour cereals $\rightarrow$ stir cereals'', 
and ``pour milk $\rightarrow$ stir cereals''. 

\textbf{Structure}:Unlike \textit{Causalogue}, 
although the entire dataset includes the 10 types of preparation 
processes of breakfasts, the number of causal structures 
far exceeds 10. Most videos do not encompass all the actions 
in a process. For example, the entire process of ``Salad" consists 
of 7 actions, but some videos are missing the ``take plate" action, 
and some videos include the actions ``cut fruit1" and ``cut fruit 2". 

\textbf{Sample}: We consider a video as a sample. The statistics of 
samples with different processes are shown as Table~\ref{tabsscd}. 

\begin{table}
   
  \caption{The number of samples in \textit{Causaction} Dataset}
  \label{tabsscd}
  \begin{center}
     \resizebox{1\linewidth}{!}{
  \begin{tabular}{c|ccccccc}
  \hline
  \multirow{2}{*}{\bf Process}&\multicolumn{7}{c}{\bf Number of Actions (Variables)}\\
  \cline{2-8}
  &4&5&6&7&8&9&all\\
  \hline
  cereals&36&-&-&-&-&-&36\\
  coffee&12&28&-&-&-&-&40\\
  friedegg&52&45&53&7&4&-&161\\
  milk&56&14&4&-&-&-&74\\
  salad&6&52&29&25&37&33&182\\
  sandwich&52&11&6&2&4&-&75\\
  tea&14&5&-&-&-&-&19\\
  pancake&-&100&26&24&33&41&224\\
  scrambledegg&8&36&30&42&33&24&173\\
  juice&65&36&24&6&-&3&134\\
  all&301&327&172&106&111&101&1118\\
  \hline
  \end{tabular}}
  \end{center}
  \end{table}

\subsubsection{Creation Process}

The original Breakfast Dataset, has annotated the frame boundaries 
of each action. Therefore, in our annotation work, 
we don’t need to ascertain which frames a causal variable contains. 
The annotators were asked to directly annotate at the action level 
to avoid inconsistencies caused by the subjectivity of watching  
videos. For example, in the coffee process, there are 6 actions, 
so a total of 15 binary relationship pairs need to be annotated. 
Specifically, we informed the annotators of the time order  
and the explanation of all actions in each process. 
After ensuring the understanding of each action, 
the annotators conducted a causal relationship evaluation 
on binary action pairs $(A, B)$ that satisfy the linear order 
relation, where $1$ signifies a belief that action $A$ has a 
causal relationship with action $B$, and $0$ represents 
no such relationship. An action pair is considered to have 
a causal relationship if the following conditions are met:

\begin{itemize}
\item According to life experience, after action $A$ happens, 
action $B$ is high-probably to occur.
\item According to life experience, after action $B$ happens, 
action A is low-probably to occur.
\end{itemize}

Finally, we binarize all annotation results, that is, 
binary pairs with a mean $>0.5$ are marked as $1$, 
and those with a mean $<0.5$ are marked as $0$.

The annotators consist of 10 researchers in the field 
of causal inference (Group A) and 217 deep-learning researchers 
(Group B). Initially, we asked Group A to annotate the two 
simplest processes, ``milk" and ``coffee," and considered 
their annotation results as the gold standard. 
Members of Group B first annotated ``milk" and ``coffee," 
with only those members having $>80\%$ consistency with Group A 
deemed qualified. In the end, 190 qualified members 
were confirmed in Group B, joining the 10 members in Group A to 
form Group C (total 200 members). Group C annotated the 
remaining 8 processes, and the statistical results 
after binarization were used as the final labels. 
During the annotation process, the consistency was $94.13\%$ 
for Group A, $88.46\%$ for the qualified members of Group B, 
and $88.74\%$ for Group C. 

\subsection{Task Definition} 
Followed by Definition~\ref{defcausalmodel}, the causal variable 
$x_{s,m,n}$ is the $n^{th}$ variable of a sample $X_{s,m}=(x_{s,m,1}, x_{s,m,2},\dots,x_{s,m,N})$. 
The causal pair $(x_{s,m,i},x_{s,m,j})$ 
($x_{s,m,i} \preccurlyeq_{X_{s,m}} x_{s,m,j}$) represents the 
causal relationship from $x_{s,m,j}$ to $x_{s,m,i}$, and the causal 
representation $\hat{x}_{s,m,i}$ represents the learned-well deep 
representation meeting the Definition~\ref{defcr}. 

Thus, in a given sample, the fundamental task of Indefinite Data is 
to \textbf{extract all causal pairs} and 
\textbf{output causal representation}  
$\hat{X}_{s,m}=(\hat{x}_{s,m,1}, \hat{x}_{s,m,2},\dots,\hat{x}_{s,m,N})$. 

\section{Experiments}
Indefinite Data is regarded as a task that outputs  
both causal structures and causal representations, 
for which we have separately evaluated causal structures 
and causal representations. Considering that our 
baseline model has a disentanglement extension, 
we have additionally designed a synthetic dataset 
to analyze the ability to estimate the effects of confounding. 

\subsection{Existing Approaches and Details of Implementation}
To the best of our knowledge, no existing approach can be  
applicable in Indefinite data. So we choose 
the SOTA work in Causal Discovery from multi-structure data and 
multi-value representation, respectively. 

In multi-structure data, we evaluate our model with ACD~\cite{lowe2022amortized} 
and AVICI~\cite{lorch2022amortized}. In multi-value representation, 
we evaluate our model with CAE~\cite{chen2023affective}, CVAE~\cite{chen2023learning}, 
and DAG-GNN~\cite{yu2019dag}. Meanwhile, for the disentanglement, 
we evaluate our model with some SOTA work focusing on latent 
confounders: \textbf{pcss}~\cite{agrawal2021decamfounder}, 
\textbf{LFCM}~\cite{squires2022causal}, and \textbf{GIN}~\cite{xie2020generalized}. 
In the experiment, we made some necessary modifications 
to these methods to adapt them to Indefinite Data. 
For example, for ACD and AVICI, we increased the 
dimensions of the hidden layers to enlarge the 
representation space, while mapping the reconstruction loss 
into the correlation relationship space. For those methods 
focusing on multi-value data, we replaced the latent 
variables with causal strength. 

In our Experiments, we utilized RoBERTa-base~\footnote{\url{https://huggingface.co/roberta-base}}
as our pre-trained model for generating word embeddings 
in the \textit{Causalogue}. Throughout the training process, 
a learning rate of 1e-5 was set, with the batch size and 
epochs set to 16 and 50, respectively. The dimension of 
the hidden layers within the network was also set to 768. 
For the \textit{Causaction}, we use less batch size with 4 to 
overcome the variable length and adopt more dimensions of 
the hidden layers (1024) to match the more complex information 
in video representation. The entire training procedure 
was conducted on a NVIDIA GEFORCE 970 RTX 3090 graphics 
processing unit. In both datasets, the 100 samples were randomly selected for 
valid set and 200 samples were randomly selected for test set. 
Each result is evaluated by 10-fold cross-validation. 

\subsection{Causal Structure}\label{secexpcs}
We evaluated the performance of recovering causal structures 
(causal graphs) on \textit{Causalogue} and \textit{Causaction}, 
using 3 different metrics: area under a receiver operating 
characteristic (AUROC), mean Squared error (MSE), 
and Hamming distance (HD). 

\begin{table*}
   
  \caption{Performance of recovering causal graphs, 95$\%$ confidence interval shown.}
  \label{tabporcg}
  \begin{center}
     \resizebox{0.8\linewidth}{!}{
  \begin{tabular}{c|cccccc}
  \hline
  \multirow{2}{*}{\bf Method}&\multicolumn{3}{c}{\bf \textit{Causalogue}}&\multicolumn{3}{c}{\bf \textit{Causaction}}\\
  \cline{2-7}
  &AUROC&MSE&HD&AUROC&MSE&HD\\
  \hline
  ACD &0.55$_{\pm 0.024}$&0.31$_{\pm 0.005}$&0.82$_{\pm 0.018}$&0.65$_{\pm 0.011}$&0.41$_{\pm 0.013}$&1.4$_{\pm 0.023}$\\
  AVICI&0.57$_{\pm 0.019}$&0.37$_{\pm 0.003}$&0.86$_{\pm 0.024}$&0.69$_{\pm 0.009}$&0.44$_{\pm 0.011}$&1.2$_{\pm 0.016}$\\
  CAE& 0.54$_{\pm 0.021}$&0.41$_{\pm 0.005}$&0.79$_{\pm 0.021}$&0.61$_{\pm 0.012}$&0.48$_{\pm 0.012}$&1.3$_{\pm 0.019}$\\
  CVAE& 0.56$_{\pm 0.014}$&0.40$_{\pm 0.001}$&0.88$_{\pm 0.013}$&0.59$_{\pm 0.011}$&0.51$_{\pm 0.015}$&1.6$_{\pm 0.021}$\\
  DAG-GNN&0.41$_{\pm 0.034}$&0.36$_{\pm 0.003}$&0.78$_{\pm 0.015}$&0.59$_{\pm 0.007}$&0.45$_{\pm 0.009}$&1.8$_{\pm 0.020}$\\
  \hline 
  Ours&\bf0.69$_{\pm 0.019}$&\bf0.26$_{\pm 0.002}$&\bf0.49$_{\pm 0.019}$&\bf0.78$_{\pm 0.008}$&\bf0.30$_{\pm 0.009}$&\bf1.1$_{\pm 0.023}$\\
  \hline
  \end{tabular}}
  \end{center}
  \end{table*}

Table~\ref{tabporcg} shows that our baseline model 
significantly outperforms existing methods with  
the applied necessary modifications. We believe this is due to 
the excessive specific assumptions made by existing methods 
for certain forms of data, which hinder their extension 
to a broader range of data forms. For instance, with DAG-GNN, 
even though we modified latent variables to adapt to 
Indefinite Data, with the acyclic constraint 
from NOTEARS~\cite{zheng2018dags}, a unique phenomenon emerges 
during the optimization process: the adjacency matrix $A$ 
tends to make $A_{ij}$ and $A_{ji}$ identical. 
This is advantageous for traditional causal data with 
unknown causal order, but conflicts with the linear order 
in Indefinite Data. Moreover, we found that methods for 
multi-structured data (ACD and AVICI) perform only second best 
to our method. This confirms that structure and representation 
are two individual aspects: multi-structure data have 
common laws, regardless of whether they are  
in single- or multi-value representation. 

In addition, the ability to generalize out of distributions 
is essential for multi-structure data. To test whether 
these models can maintain robustness when encountering 
new causal structures, we designed a simple 10-fold experiment. 
In each fold, we randomly selected 2 structures 
(of \textit{Causalogue}) or processes (of \textit{Causaction}) 
for the test set, with all its samples prohibited 
from appearing in the train and valid sets. 

\begin{table*}
   
  \caption{Performance of recovering causal graphs out of distribution}
  \label{tabpfrcgofd}
  \begin{center}
     \resizebox{0.8\linewidth}{!}{
  \begin{tabular}{c|cccccc}
  \hline
  \multirow{2}{*}{\bf Method}&\multicolumn{3}{c}{\bf \textit{Causalogue}}&\multicolumn{3}{c}{\bf \textit{Causaction}}\\
  \cline{2-7}
  &AUROC&MSE&HD&AUROC&MSE&HD\\
  \hline
  ACD &0.51$_{\pm 0.031}$&0.46$_{\pm 0.025}$&1.63$_{\pm 0.049}$&0.53$_{\pm 0.034}$&0.58$_{\pm 0.028}$&2.1$_{\pm 0.046}$\\
  AVICI&0.51$_{\pm 0.045}$&0.46$_{\pm 0.032}$&1.13$_{\pm 0.051}$&0.60$_{\pm 0.049}$&0.54$_{\pm 0.037}$&2.8$_{\pm 0.041}$\\
  CAE& 0.46$_{\pm 0.033}$&0.46$_{\pm 0.035}$&1.37$_{\pm 0.054}$&0.61$_{\pm 0.035}$&0.63$_{\pm 0.039}$&2.5$_{\pm 0.055}$\\
  CVAE& 0.49$_{\pm 0.044}$&0.48$_{\pm 0.028}$&1.37$_{\pm 0.046}$&0.52$_{\pm 0.039}$&0.49$_{\pm 0.045}$&2.9$_{\pm 0.048}$\\
  DAG-GNN&0.33$_{\pm 0.041}$&0.43$_{\pm 0.029}$&1.45$_{\pm 0.049}$&0.48$_{\pm 0.044}$&0.53$_{\pm 0.039}$&2.7$_{\pm 0.043}$\\
  \hline 
  Ours&\bf0.61$_{\pm 0.024}$&\bf0.35$_{\pm 0.008}$&\bf0.94$_{\pm 0.027}$&\bf0.66$_{\pm 0.016}$&\bf0.42$_{\pm 0.014}$&\bf1.8$_{\pm 0.031}$\\
  \hline
  \end{tabular}}
  \end{center}
  \end{table*}

Table~\ref{tabpfrcgofd} records the results of the 
cross-distribution testset. Our method consistently 
outperforms existing methods, and the entire statistical 
result shows a situation similar to that of Table~\ref{tabporcg}. 
Additionally, we noticed that the standard deviation 
of our method is much lower than other methods. 
We consider that for the reason there are 
similarities among some structures in the dataset. 
For instance, in the \textit{Causalogue} dataset, 
Hybrid$\_$II is very similar to Hybrid$\_$I, 
but significantly different from the other 8 structures. 
In the \textit{Causaction} dataset, many common causal 
relationships exist among ``friedegg'' and ``pancake''. 
When these structures are chosen for the test set in certain folds, 
the model can find ``answers" from similar structures 
in the train set. However, when similar structures are all present 
in the test set fold (e.g., the test set includes Hybrid$\_$I 
and Hybrid$\_$II), it is difficult for the trained structures 
to manifest apparent invariance. However, the lowest 
standard deviation once-again demonstrated the superiority 
of our baseline in releasing many assumptions about data forms. 
In other words, existing methods tend to rely on 
specific hypotheses to recover causal relationships, 
while our approach is more inclined to let the model 
itself learn the causal relationships. 

\subsection{Causal Representation}
Evaluating causal representation is another crucial aspect 
of Indefinite Data. According to Definition~\ref{defcr}, 
causal representation needs to be evaluated on both 
correlation and causation. Specifically, we assume 
$x_{i}$ and $x_j$ to be any two causal representations 
that need to be tested. We propose a correlation matrix, 
$Cor$, to verify the performance in correlation, 
where $Cor_{ij} = cossin(x_i, x_j)$. Moreover, 
we train a downstream linear layer to extract causal relationships, 
$Cas_{ij} = linear layer (x_i || x_j)$. 
Both $Cor$ and $Cas$ are evaluated by AUROC and MSE, 
to demonstrate the performance of the causal representation 
in correlation and causation, respectively. 

\begin{table*}
   
  \caption{Performance of learning causal representations}
  \label{tabpolcr}
  \begin{center}
     \resizebox{1\linewidth}{!}{
  \begin{tabular}{c|cccccccc}
  \hline
  \multirow{3}{*}{\bf Method}&\multicolumn{4}{c}{\bf \textit{Causalogue}}&\multicolumn{4}{c}{\bf \textit{Causaction}}\\
  \cline{2-9}
  &\multicolumn{2}{c}{\bf $Cas$}&\multicolumn{2}{c}{\bf $Cor$}&\multicolumn{2}{c}{\bf $Cas$}&\multicolumn{2}{c}{\bf $Cor$}\\
  &AUROC&MSE&AUROC&MSE&AUROC&MSE&AUROC&MSE\\
  \hline
  ACD   & 0.52 $_{\pm 0.026}$& 0.64 $_{\pm 0.074}$& 0.91 $_{\pm 0.013}$& 0.43 $_{\pm 0.022}$& 0.59$_{\pm 0.006}$ & 0.39 $_{\pm 0.009}$& 0.88 $_{\pm 0.001}$& 0.28 $_{\pm 0.005}$\\
  AVICI  & 0.57 $_{\pm 0.021}$& 0.59 $_{\pm 0.032}$& 0.91 $_{\pm 0.017}$& 0.31 $_{\pm 0.016}$& 0.62 $_{\pm 0.002}$& 0.34 $_{\pm 0.009}$& 0.94$_{\pm 0.001}$&0.21 $_{\pm 0.001}$ \\
  CAE    & 0.61 $_{\pm 0.023}$& 0.52  $_{\pm 0.047}$& 0.93 $_{\pm 0.011}$& 0.32 $_{\pm 0.013}$& 0.64 $_{\pm 0.001}$& 0.36 $_{\pm 0.011}$& 0.92 $_{\pm 0.003}$&0.25$_{\pm 0.003}$\\
  CVAE   & 0.62 $_{\pm 0.021}$& 0.55  $_{\pm 0.066}$& 0.91 $_{\pm 0.006}$& 0.29 $_{\pm 0.024}$& 0.61 $_{\pm 0.005}$& 0.31 $_{\pm 0.005}$& 0.92$_{\pm 0.001}$ &0.23$_{\pm 0.003}$\\
  DAG-GNN& 0.59 $_{\pm 0.019}$& 0.55 $_{\pm 0.059}$& 0.90 $_{\pm 0.017}$& 0.39 $_{\pm 0.009}$& 0.63 $_{\pm 0.003}$& 0.33$_{\pm 0.013}$ & 0.91 $_{\pm 0.002}$&0.26 $_{\pm 0.002}$ \\
  \hline 
  Ours& \bf0.68$_{\pm 0.016}$ & \bf0.43$_{\pm 0.058}$ & \bf0.95$_{\pm 0.008}$ & \bf0.26$_{\pm 0.011}$ & \bf0.79$_{\pm 0.005}$  & \bf0.26$_{\pm 0.004}$ & \bf0.96$_{\pm 0.002}$ & \bf0.15$_{\pm 0.001}$ \\
  \hline
  \end{tabular}}
  \end{center}
  \end{table*}

Table~\ref{tabpolcr} presents the performance 
in correlation and causation. The results suggest that 
the representation more easily grasps the information 
of correlation, while causation, an asymmetric and underlying  
relation, poses a more challenging topic in representation 
learning. Moreover, our method significantly outperforms others, 
even when we have modified them to adapt multi-value 
representations. This reason aligns with Section~\ref{secexpcs}, 
for instance, the causality in ACD is based on the 
Granger causality hypothesis in time series, which stresses 
the faithfulness of single clues to causation. 
However, when it is expanded to other types of data 
(like the current Indefinite Data), it is tough to ascertain 
the correct set of parent nodes for causal representation. 
In addition, similar to the performance of causal structure, 
methods of multivalued representation (CAE, CVAE, DAG-GNN) 
also show superior performance in Table~\ref{tabpolcr} 
over the multi-structure data methods. Thus, we can 
emphasize that representation and structure are two separate 
dimensions, and the concurrent existence of multi-value 
representation and multi-structure data leads to new challenges.

\subsection{Disentanglement}
\begin{figure*}
  \centering
  \subfloat[$\#$Sample]{
    \includegraphics[width=0.23\linewidth]{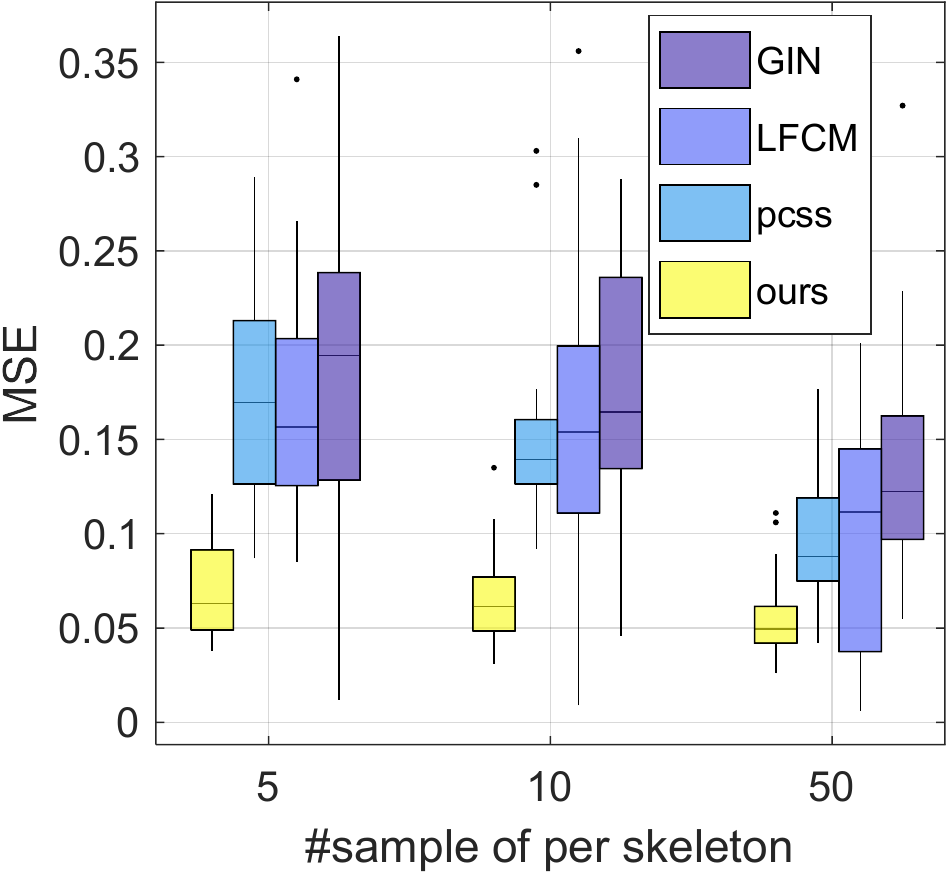}}
  \subfloat[$\#$Confounder]{
    \includegraphics[width=0.23\linewidth]{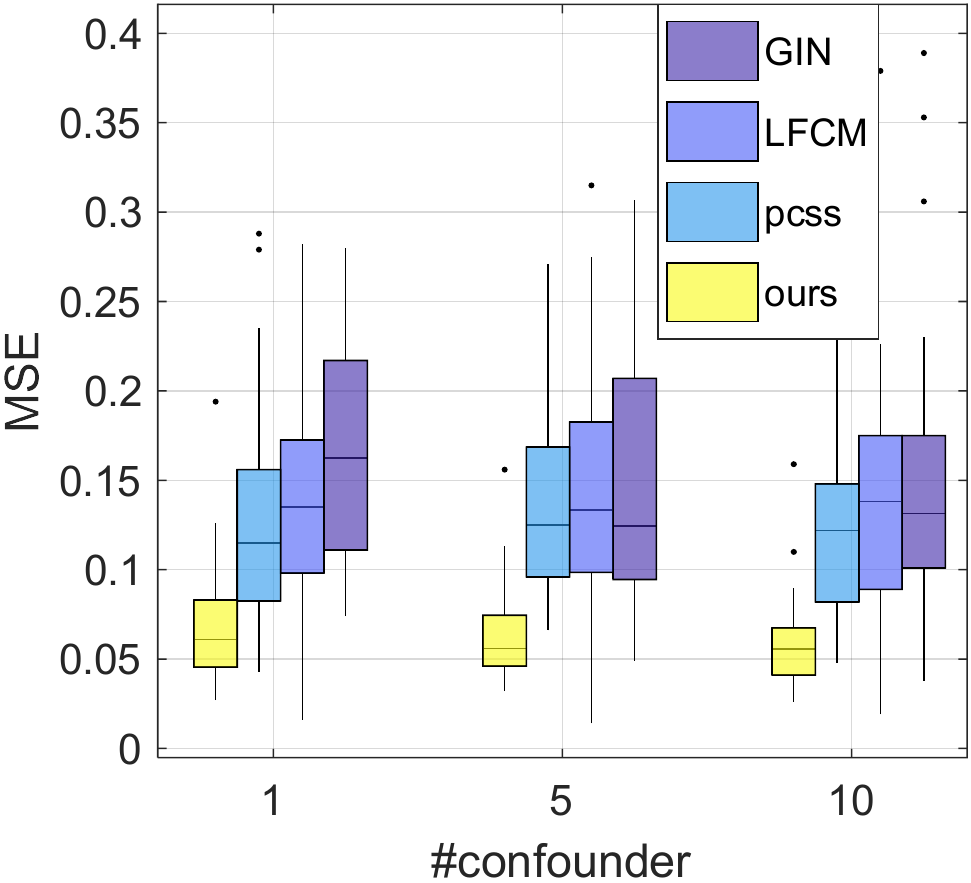}}
  \subfloat[$\#$Observed nodes]{
    \includegraphics[width=0.23\linewidth]{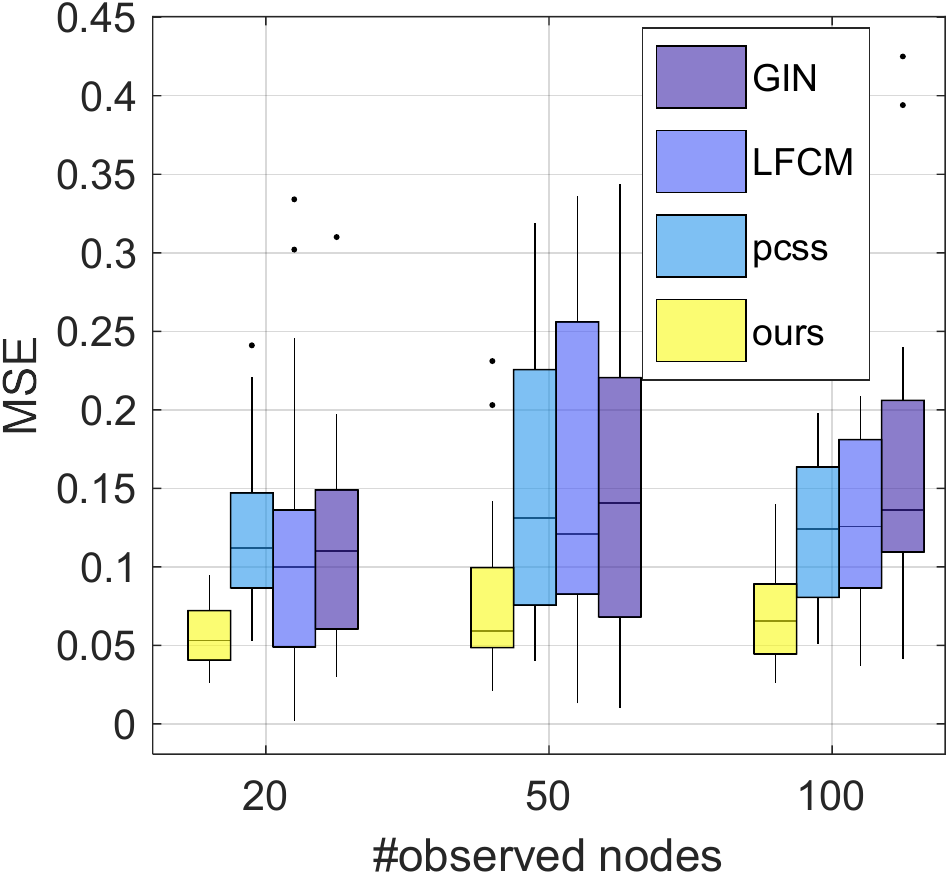}}
  \subfloat[Pervasiveness]{
    \includegraphics[width=0.23\linewidth]{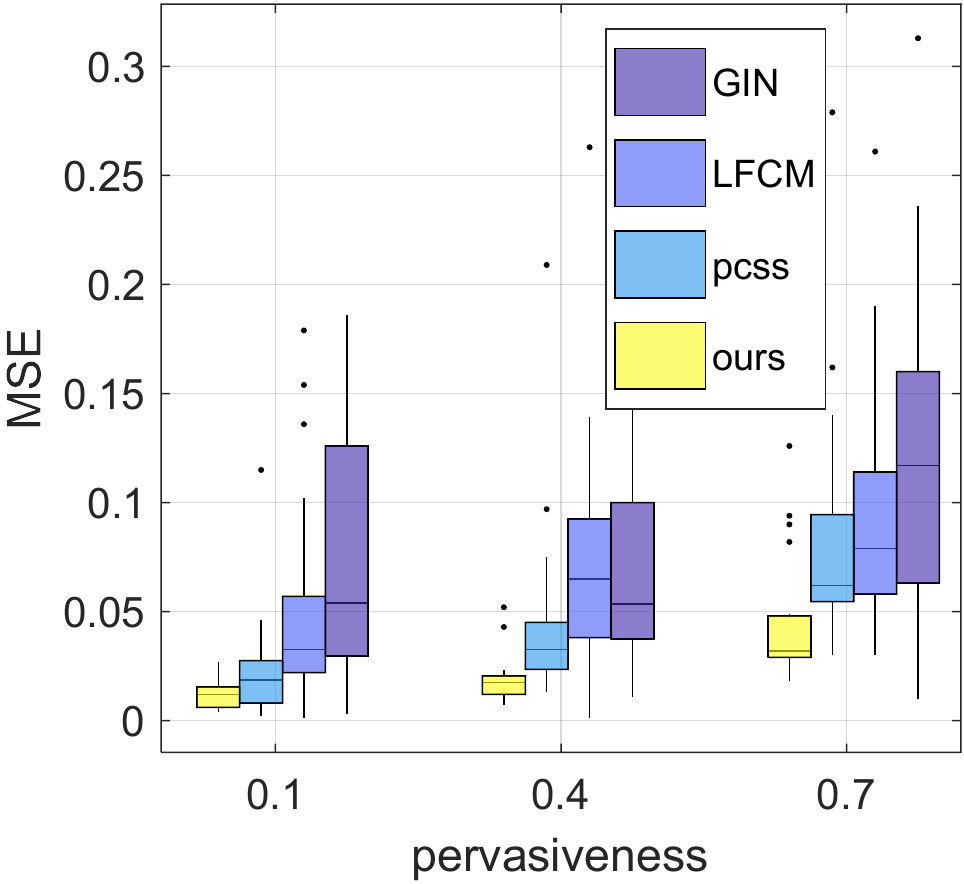}}
  \caption{MSE error across all ingredients setting for estimating 
  $C$ via GIN, LFCM, pcss, and ours.}
  \label{figmse}
\end{figure*} 

We created a set of synthetic dataset to evaluate the estimation 
of confounding effects. Specifically, We randomly draw Causal DAG 
from a random graph model with an expected neighborhood size of 5 and 
consider graphs with the number of observed nodes $N\in \{20, 50, 100\}$. 
For probing how our approach is affected by the pervasiveness of 
confounding, we assume that each confounder $l_{k}$ is a direct cause 
of node $x_{i}$ with a chance $P \in \{0.1, 0.4, 0.7\}$. Given the graph, 
we stochastically set a trend type for each causal strength weight 
with equal probability. Meanwhile, we add $N(0,\sigma^{2}_{noise})$ 
noise to each node. Finally, we consider the number of confounders 
$K \in \{1, 5, 10\}$ and the number of samples of each skeleton 
$n \in \{5, 10, 50\}$, respectively.

In Figure~\ref{figmse}, we quantify the mean-squared estimation 
(MSE) error of $C$. Our method likewise performs best in 
all ingredient settings, demonstrating that 
our confounding disentanglement pool the statistical strength 
better than other estimation algorithms in multi skeleton data. 
Besides, combined with the conclusion in
~\cite{agrawal2021decamfounder}, this error should decrease as 
the number of samples $n$ increases. Figure~\ref{figmse} (a) is 
exactly indicative of this conclusion. 

\section{Discussion}

This paper focuses on causal inference for a novel paradigm 
of data - Indefinite Data, characterized by multi-structure 
data and multi-value representations. These two features 
differ greatly from traditional experimental data, 
thereby causing existing methods to be non-adaptive 
to Indefinite Data. To provide a good starting point 
for causal research on Indefinite Data, we introduce 
two brand new datasets- \textit{Causalogue} and \textit{Causaction}, 
analyze the challenges brought about by the coexistence 
of multi-structure data and multi-value representations, 
and propose a corresponding probabilistic framework. 
In the experiments, we exhibit benchmark results 
for both structure and representation and 
share intrinsic insights in the extension of disentanglement. 

However, such a probabilistic framework is inconsistent 
for the learning of causal structure and causal representation. 
Specifically,  for multi-structure scenarios, 
learned causal strength $\hat{f}=h_1(X, \varphi)$ 
($\varphi$ represents the parameters of encoder), 
and for multi-value representation, $\hat{X}=h_2(X, \hat{f})$. 
And $D>1$ makes $loss(\hat{X}, X)$ be replaced with 
$loss(cs(\hat{X}), cs(X))$. 
Therefore, when only $loss(\hat{f}, f)$ exists, 
we can get $\hat{f}=f$and $f \Leftrightarrow X$. However, 
we cannot guarantee $\hat{X}\Leftrightarrow \hat{f}$or $X=\hat{X}$, 
thus severe inconsistencies exist. 
This phenomenon doesn't exist in other data paradigms, 
For instance, when $M=1 \& D>1$, we can guarantee 
$\hat{f} \Leftrightarrow \hat{X}$, and thus 
ensure $X=\hat{X}$ by a fixed causal structure. 
Meanwhile, when $M>1 \& D=1$, 
$\hat{f}=f\Leftrightarrow X=\hat{X}$ can be achieved through 
$loss(\hat{X}, X)$ and $loss(\hat{f}, f)$. 
We have elaborated on this problem and proposed an 
intervention-based improvement in our latest work~\cite{chen2023ssl}.

Causal research on Indefinite Data will be a long-term topic. 
The ultimate goal of multi-structure data is to exhibit 
cross-distribution learning capability, while multi-value 
representations will eventually rely on a model-driven 
learning process. In previous work, we posited such a 
challenge, and this paper makes it feasible from a set of research 
basis. We hope this will attract an increasing number of 
researchers to contribute to the growing body of work 
applying causal inference to the real world.

\bibliographystyle{IEEEtran}
\bibliography{tpami}

\end{document}